\documentclass[runningheads]{llncs}
\usepackage[T1]{fontenc}
\usepackage{graphicx}
\usepackage{amsmath}
\usepackage{amssymb}
\usepackage{algorithm}
\usepackage{setspace}
\usepackage{algpseudocode}
\usepackage{booktabs}
\usepackage{multirow}
\usepackage{appendix}
\usepackage{subcaption}
\usepackage{hyperref}
\usepackage{subcaption}
\usepackage{color}
\begin{document}

\title{Balancing Knowledge Distillation for Imbalance Learning with Bilevel Optimization}

\titlerunning{Balancing Knowledge Distillation for Imbalance Learning}
\author{Anh B.H. Nguyen$^{\star}$ \and
Ba Tho Phan\thanks{Equal contribution.} \and
Viet Cuong Ta\thanks{Corresponding author.}}
\authorrunning{Anh Nguyen, BT Phan, VC Ta}
\institute{VNU University of Engineering and Technology, Hanoi, Vietnam
\email{\{22021100,23020163,cuongtv\}@vnu.edu.vn}}
\maketitle

\begin{abstract}
Knowledge distillation transfers knowledge from a high capacity teacher to a compact student using a mixture of hard and soft losses. On imbalanced data, a fixed weighting between hard and soft losses becomes brittle the learning process. Recent studies try to reweight these components in long-tailed settings. However, most of these methods do not adapt weights at the sample-wise level and do not take into account the student’s behavior during training. 
To address this, we propose BiKD - a bilevel framework that dynamically balances hard and soft losses for each sample. We employ a weight generation network that produces adaptive per-sample weights, guided by a small balanced validation set. The student is now trained with an unconstrained combination of weighted hard and soft losses, allowing the student to relax both terms. We further propose a multi-step SGD strategy to optimize the weight model more accurately and efficiently. Experiments on long-tailed CIFAR‑10/100 show that our approach surpasses recent balanced distillation methods across imbalance factors.
Our implementation is available at https://github.com/phan-tho/Bilevel-balancing-kd.

\end{abstract} 
\section{Introduction}

Deep learning models have achieved notable performance in the fields of computer vision with various applications. However, the large model architectures play a significant role in traditional training procedures. The advent of large models poses a significant challenge for deployment on edge devices like mobile phones due to their heavy storage and computational requirements. To address this, Knowledge Distillation (KD)~\cite{KD} is one of the prominent techniques that use a teacher–student strategy. In KD, a lightweight student network is trained using a combination of teacher knowledge and label supervision.

However, typical setups in KD methods~\cite{KDsurvey} implicitly assume that the transfer set, which is used to train both teacher and student models, is roughly class-balanced. In practice, many real-world distributions are long-tailed~\cite{survey_imbalanced}, a few head classes account for most samples, while tail categories are scarce. This causes both the high-capacity teacher and the student to degrade in performance on tail classes. ~\cite{LFME,BKD}.
Under such prior shift, using a fixed weight for the label-supervised (hard) loss and the teacher-guided (soft) loss is brittle and can bias learning toward head classes.

For this reason, recent studies~\cite{DiVE,LFME,BKD} have explored knowledge distillation in the presence of biased teachers and imbalanced transfer sets. A prominent line of methods~\cite{he2024joint,BKD} assigns appropriate weights to the hard and soft losses to balance label supervision and teacher guidance. 
Despite outperforming distillation methods in imbalanced settings, these weighting based KD approaches still face two major limitations. First, it uses a single soft-loss weight which is shared by all samples in a class, ignoring the feature patterns. This contrasts with recent studies~\cite{RWKD,WSL} which emphasize that the weight of hard and soft losses should vary at the sample level. Second, the weights are not adapted to the student’s behavior. This is not appropriate in the context of imbalance learning. 

In this work, we propose a BiKD framework that frames KD on an imbalanced transfer set as a bilevel problem. Within this framework, we can train the student on the imbalanced data with adaptive weights. We use a small network as a meta model that learns the mapping to the output weights, supervised by the student’s performance on the unbiased set. To be more accurate and efficient, we use a one-step virtual update and accumulate hypergradients to update the meta model. This design preserves the benefits of multi-step optimization without storing full student updating computation graphs. Experiments on long-tailed CIFAR‑10/100 show that our approach achieves the most stable performance in comparison to other distillation methods.

%
\section{Related work}

Knowledge Distillation aims to transfer the predictive behavior of a large teacher network into a smaller student model. The main idea is that the student model mimics the teacher model in order to obtain a competitive or even a superior performance. Follow-up works distill richer information such as intermediate features~\cite{Heo2019ACO,FitNet}, spatial attention maps~\cite{guo2023classattentiontransferbased,Zagoruyko2016PayingMA}, or relational structure among samples~\cite{Park2019RelationalKD,Tung2019SimilarityPreservingKD}, and KD~\cite{KD} is now widely used for model compression and regularization. The standard objective combines the cross-entropy loss on ground-truth labels with a Kullback–Leibler divergence between the student and teacher distributions. More recently, several methods argue that the trade-off between hard and soft losses should vary across samples. WSL~\cite{WSL} rethinks soft labels from a bias–variance perspective and proposes weighted soft labels to achieve a better sample-wise balance. RW-KD~\cite{RWKD} uses a meta learning framework to adapt weights for the hard and soft losses per sample based on gradient signals. However, these methods typically assume roughly balanced training data.

Long-tailed recognition methods show that standard training strongly favors head classes when class frequencies are highly skewed~\cite{Liu2019LargeScaleLR}. In knowledge distillation, when a teacher is trained on the long-tailed data, naive KD transfers this teacher bias and can further degrade tail-class performance of the student. To address this, recent balanced KD methods~\cite{he2024joint,BKD} introduce class-balanced reweighting on the distillation term so that soft targets emphasize minority classes without losing head-class information. Class-balanced or class-conditional KD methods~\cite{CBD,CCKD} further decouple representation learning and distillation, then align teacher and student class-conditional features to mitigate class imbalance. Multi-expert distillation approaches~\cite{SHIKE,LFME} build several teachers on less imbalanced subsets and aggregate them into a single student tailored to long-tailed data. DiVE~\cite{DiVE} tackles long-tailed recognition by interpreting teacher predictions as virtual examples and explicitly flattening this virtual example distribution before distillation.

\section{Methodology}

\subsection{Problem  setup}
Given a set of imbalance training data $\mathcal{D}_{\text{train}} = \{(x_i, y_i)\}^N_{i=1}$ with one-hot labels $y_i\in\{0,1\}^C$. $\mathcal{D}_{\text{train}}$ follows the standard long-tailed setting~\cite{imbalancedcifar}, where the number of training samples per class decreases exponentially from head to tail. Let $T(\cdot)$ be an imbalanced pretrained \textbf{teacher} on $\mathcal{D}_{\text{train}}$ and $S_\theta(\cdot)$ a \textbf{student} with parameters $\theta$.
Following~\cite{KD}, KD transfers “soft” targets from the teacher at a temperature $\tau$.
The student learns from a \emph{hard} supervised loss and a \emph{soft} distillation loss:
\[
\mathcal{L}_{\text{train}}(\theta)
=\frac{1}{N}\sum_{i=1}^N \Big[
(1-\alpha)\,\mathrm{CE}\!\big(y_i,\,p_S(x_i)\big)
+\alpha\,\tau^2\,\mathrm{KL}\!\big(p_T^\tau(x_i)\,\|\,p_S^\tau(x_i)\big)
\Big],
\]
where $\alpha\in[0,1]$ mixes the two terms and $\tau^2$ keeps the soft term’s gradients comparable when $\tau>1$. $p_S^\tau$ and $p_T^\tau$ are soft predictions with a temperature $\tau$ of the student and teacher, respectively.
Since the teacher is fixed, $\mathrm{KL}(p_T^\tau\|p_S^\tau)$ is equivalent to cross-entropy with soft targets $p_T^\tau$.
For clarity, we denote
$\mathcal{L}_{\text{hard}}(x_i)=\mathrm{CE}(y_i,p_S(x_i))$ and
$\mathcal{L}_{\text{soft}}(x_i)=\tau^2\,\mathrm{KL}(p_T^\tau(x_i)\|p_S^\tau(x_i))$.

It is worth pointing out that accurately estimating $\alpha$ is often difficult. The coefficient between hard and soft loss depends on many factors, such as sample inputs~\cite{CCKD,WSL} or gradient signals during training~\cite{RWKD,Safaryan2023KnowledgeDP}. This hyperparameter exerts an even stronger training influence when the transfer set $\mathcal{D}_{\text{train}}$ is imbalance. In the long-tailed setting, relying on constant $\alpha$ prevents the student from adapting their relative weights during training. It cannot simultaneously strengthen both gradient signals when both are informative or suppress both in contrast. These limitations suggest that hard and soft weights should be adaptive and not restricted to sum to one.

Inspired by~\cite{RWKD}, we first introduce a bilevel framework to automatically adapt the weights between the hard and soft losses for each sample $x_i$. Within this framework, the coefficients are optimized and continuously updated during training to achieve the best performance of the student model.
\subsection{Sample-wise adaptive weights through bilevel framework}
In our framework, we assume that there is a validation set of unbiased data examples $\mathcal{D}_{\text{val}} = \{(x_j, y_j)\}^M_{j=1}$ with $M$ much smaller than $N$. To best balance the information carried by the hard loss and the soft loss, we propose to construct a weight network, serving as a meta model. This model takes as input a pair of per-sample cross-entropy losses and attempts to produce the weights for the hard and soft losses for this example. The meta model is parameterized as a function with parameter $\phi$ as follows:
\[ \left[w_i^{\text{hard}}, w_i^{\text{soft}}\right] = f_\phi\left(\mathrm{CE}(y_i, p_T( x_i)), \mathrm{CE}(y_i, p_S( x_i))\right),
\]
This pair cross-entropy losses summarizes how both models fit the ground-truth label on $x_i$, provide a compact description of the training signal for this sample. WSL \cite{WSL} also applies a function of two losses to reweight the soft loss. In contrast, our learnable meta model receives the same inputs but tries to approximate the mapping from these loss pairs to $w_i^\text{hard}, w_i^\text{soft}$. The goal of $f_{\phi}$ is to determine how strongly the student should use each loss term for a given sample.\\
The student model $S_\theta$, which we aim to train and use for prediction, is trained by the proposed adaptive training loss: 
\begin{equation}
\mathcal{L}_{\text{train}}(\theta, \phi) = \frac{1}{N} \sum_{i=1}^{N} \left[ w_i^{\text{hard}}\ \mathcal{L}_{\text{hard}}(x_i) + w_i^{\text{soft}} \mathcal{L}_{\text{soft}}(x_i) \right ]
\label{trainingloss}
\end{equation}
Equipped with this training loss, the total weight on the hard and soft terms per-sample is not constrained to sum to one. This relaxation increases the student’s flexibility when distilling under class imbalance. For example, for samples $x_i$ belonging to head classes that the student can handle and distill effectively, both $w_i^{\text{hard}}$ and $w_i^{\text{soft}}$ should be small simultaneously to avoid negative impacts on the student generalization.\\
The optimal selection of the student $\theta$ is based on its validation performance: \begin{equation}
\mathcal{L}_{\text{val}}(\theta) = \frac{1}{M} \sum_{j\in \mathcal{D}_{\text{val}}} \mathcal{L}_{\text{hard}}(x_j) \label{eq:val_loss} \end{equation}
Without explicitly linking the meta model and the student, there is no way to ensure that: First, the student model $S_\theta$ achieves effective distillation knowledge from the teacher on an imbalanced transfer set, if the weights provided by the meta model do not align with the ground-truth optimal weights. Second, that the generated weights for each example are indeed meaningful, since directly training $f_\phi$ is not possible without observing the student performance on the validation set. By these observations, we link two models together via a bilevel optimization framework. The motivation for this scheme is that if the weights generated by the meta model are high quality, then a student model trained with such weighted losses as supervision should achieve low loss on the validation set. 
Formally, this framework can be formulated as the following optimization problem: 
\[
\min_\phi \mathcal{L}_{\text{val}}(\theta^*(\phi)) \hspace{0.5cm} s.t. \hspace{0.5cm} \theta^*(\phi) = \arg \min_\theta \mathcal{L}_{\text{train}}(\theta, \phi)
\]
\begin{figure}[!t]
    \centering
    \includegraphics[width=1\linewidth]{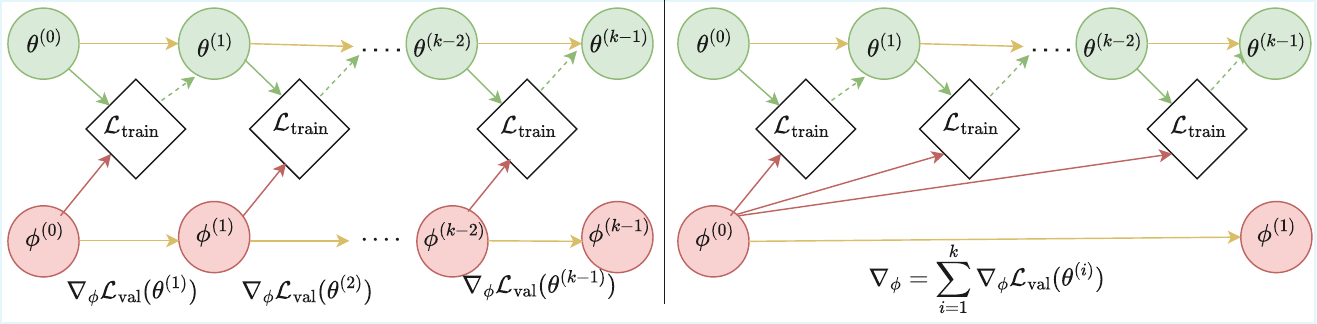}
    \caption{This figure illustrates the difference between the online optimization and our proposed strategy. 
    } 
    \label{fig:k_steps}
\end{figure}
Here, $\theta^{*}(\phi)$ denotes the student parameters obtained by training on $\mathcal{D}_{\text{train}}$ with weights generated by $f_\phi$.
The meta model is then updated by the hypergradient  $\nabla_{\phi}\,\mathcal{L}_{\text{val}}\!\big(\theta^{*}(\phi)\big)$, which reflects the student performance on the validation set.\\
We observe that the loss $\mathcal{L}_{\text{val}}(\theta^*(\phi))$ depends only implicitly on the meta parameter $\phi$ via the trained student $\theta^*(\phi)$. Therefore, accurately approximating $\theta^*(\phi)$ is necessary to reduce the noisy estimation. In this paper, we propose to employ a multiple SGD steps update to estimate the optimal student $\theta^*$. Specifically, we update the student parameters with the current weighted training loss, each step update can be formalized by:

\begin{equation}
\theta^{(t+1)} = \theta^{(t)} - \eta_\theta \nabla_\theta \mathcal{L}_{\text{train}}(\theta^{(t)}, \phi^{(t)})    \label{eq:student_update}
\end{equation}
$\eta_\theta$ denotes the learning rate of the student optimizer. We refer to this process as the inner loops. After updating $k$ steps of $\theta$, the outer loop updates the meta parameters $\phi$ by evaluating the hypergradient $\nabla_\phi \mathcal{L}_{\text{val}}(\theta^{(t+k)})$:
\begin{equation}
\phi^{(t+1)} = \phi^{(t)} - \eta_\phi \nabla_\phi \mathcal{L}_{\text{val}}(\theta^{(t+k)}) \label{eq:meta_update}
\end{equation}
$\eta_\phi$ denotes the learning rate of the meta model. In practice, directly computing the hypergradient $\nabla_\phi \mathcal{L}_{\text{val}}(\theta^{(t+k)})$ in Eq.\eqref{eq:meta_update} is inefficient, since fully unrolling $k$ inner updates would require storing $k$ computation graphs of the student $\theta$.
To mitigate this, we adopt a one-step unrolled update with a virtual student. Then we evaluate the validation objective at this virtual model and backpropagate to obtain a one-step hypergradient. We accumulate these hypergradients over inner loops to approximate $\nabla_\phi \mathcal{L}_{\text{val}}(\theta^{(t+k)})$, thereby avoiding storage of the entire inner computation graphs. Specifically, at the time step $t$, we initiate $\nabla_\phi =0$ and form a \emph{virtual student} $\theta'$ that loads the state of the student $\theta^{(t)}$. We update the virtual student by taking one gradient step on the current training minibatch:
\[
\theta'(\phi) \;=\; \theta' \;-\; \eta\,\nabla_\theta \mathcal{L}_{\text{train}}(\theta';\phi^{(t)}).
\]
We then evaluate the validation objective at $\theta'(\phi)$ and backpropagate through this virtual step to accumulate the hypergradients:
\[
\nabla_\phi = \nabla_\phi + \nabla_\phi\mathcal{L}_{\text{val}}(\theta'(\phi))
\]
The process continues for $k$ steps, we then update the meta model with information from the previous $k$ step as follows: 
\begin{align}
\phi^{(t+1)} &= \phi^{(t)} - \eta_\phi \nabla_\phi \label{eq:our_metaupdate}
\end{align}
With $k=1$, our optimization approach reduces to the online strategy~\cite{RWKD,Ren2018LearningTR,Shu2019MetaWeightNetLA}.
Figure \ref{fig:k_steps} provides an illustration of our computational graph. The detailed implementation is given in Algorithm \ref{our_algorithm}.
\begin{algorithm}[!t]
\caption{Bilevel Knowledge Distillation (BiKD)}
\label{our_algorithm}
\begin{algorithmic}[1]
\Require The teacher model $T(.)$, the student model $S_\theta(.)$, the meta model $f_\phi(.)$, 
         $\mathcal{D}_{\text{train}}$, $\mathcal{D}_{\text{val}}$,
         learning rates $\eta_\theta, \eta_\phi$, epochs $T$, inner gradient steps $k$.
\State $count = 0$, $\nabla_\phi = 0$
\For{$ t=0, ..., T - 1$}
    \For {batch $\mathcal{B}_{\text{train}} = \{(x_i, y_i)\} \subset   \mathcal{D}_{\text{train}}$}
    \State $\theta' \gets \theta^{(t)}$.
    \State $w_i^{\text{hard}}, w_i^{\text{soft}} \gets f_{\phi^{(t)}}(\mathrm{CE}(y_i, p_T( x_i)), \mathrm{CE}(y_i, p_S( x_i)))$
    \State \(\mathcal{L}_{\text{train}}(\theta', \phi^{(t)}) = \frac{1}{|\mathcal{B}_{\text{train}}|} \sum_{i}\left[ w_i^{\text{hard}}\ \mathcal{L}_{\text{hard}}(x_i) + w_i^{\text{soft}} \mathcal{L}_{\text{soft}}(x_i) \right ]
    \)

    \State $\theta'(\phi) \gets \theta'- \eta_\theta \nabla_\theta \mathcal{L}_{\text{train}} \big|_{\theta = \theta'}
    $
    \State Sample validation mini-batch $\mathcal{B}_{\text{val}} = \{(x_j^v, y_j^v)\}$ from $\mathcal{D}_{\text{val}}$.
    \State \(\mathcal{L}_{\text{val}}(\theta'(\phi)) = \frac{1}{|\mathcal{B}_{\text{val}}|} \sum_{j} \mathcal{L_{\text{hard}}}(x_j)\)
    \State $\nabla_\phi \leftarrow \nabla_\phi +  \nabla_\phi \mathcal{L}_{\text{val}}(\theta'(\phi)) \big|_{\phi= \phi^{(t)}}$
    \If {$\text{count} \ \% \ k =0$} \State \hspace{0cm} $\phi^{(t+1)} \leftarrow \phi^{(t)} - \eta_\phi \nabla_\phi$, $\nabla_\phi \gets 0$
    \Else:\State $\phi^{(t+1)} \leftarrow \phi^{(t)} $
    \EndIf
    
    \State \hspace{-0.7cm} $w_i^{\text{hard}}, w_i^{\text{soft}} \gets f_{\phi^{(t+1)}}(\mathrm{CE}(y_i, p_T( x_i)), \mathrm{CE}(y_i, p_S( x_i)))$
    \State \hspace{-0.7cm} $
    \theta^{(t+1)} \leftarrow \theta^{(t)} - \eta_\theta \nabla_\theta \mathcal{L}_{\text{train}} (\theta^{(t)}, \phi^{(t+1)}) \big|_{\theta= \theta^{(t)}}
    $
    \State \hspace{-0.7cm} $count \leftarrow count + 1$
    \EndFor
\EndFor
\end{algorithmic}
\end{algorithm}
\subsection{Overall framework}
\begin{figure}
    \centering
    \includegraphics[width=1\linewidth]{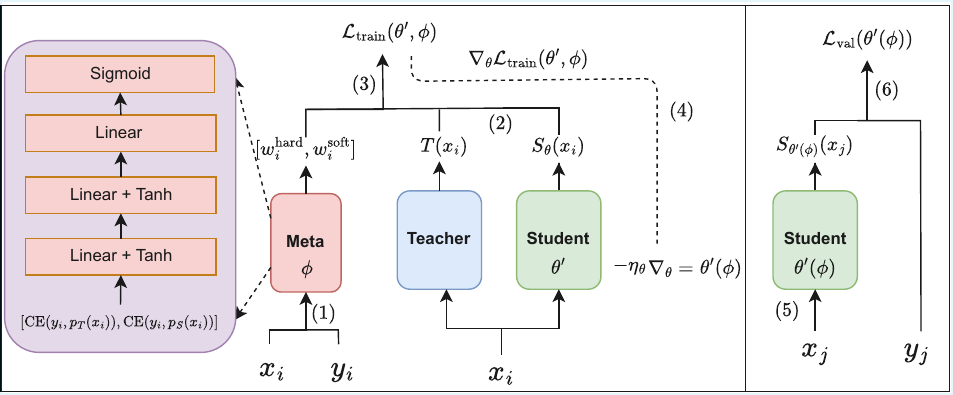}
    \caption{This figure illustrates our framework.}
    \label{computational_graph}
\end{figure}
Figure \ref{computational_graph} demonstrates our framework, the procedure is as follows: (1) Initializing the virtual student $\theta'$ and computing the weights $[w^{\text{hard}}, w^{\text{soft}}]$ from the meta model on the training data \((x_i, y_i) \sim D_{\text{train}}\); (2) Evaluating the output of the virtual student $\theta'$ and the teacher $T(.)$ on $x_i$; (3) Computing the training loss according to Eq.\eqref{trainingloss}; (4) Deriving the gradient of the training loss to update the virtual student; (5) Computing the output of the updated virtual student $\theta'(\phi)$ on the validation sample $x_j$ in $D_{\text{val}}$; and finally, (6) Evaluating the validation loss $\mathcal{L}_{\text{val}}$ to get a one-step hypergradient. After accumulating the hypergradients, we re-compute $w^{\text{hard}}$ and $w^{\text{soft}}$ to update the real student model.

\subsection{Analysis of the meta model weighting mechanism}
We detail how the meta model generates the weights for the hard and soft losses in the training objective. Under the bilevel framework, $w^{\text{hard}}$ prevents the student from biasing toward majority class samples, while $w^{\text{soft}}$ encourages the student to follow
informative teacher signals during distillation.\\
The meta parameters update in Eq.(\ref{eq:our_metaupdate}) can be implemented as:

\begin{align}
\phi^{(t+1)} = \phi^{(t)} + \dfrac{\eta_\phi \eta_\theta}{N} \sum_{x_j \in \mathcal{D}_{\text{train}}} &\big[\big( \sum^{k-1}_{i=0}g_{\text{val}}^{(i+1)} g^{(i)}_{\text{hard}}(x_j)\big)^T \dfrac{\partial w_j^{\text{hard}}}{\partial \phi} \notag \\
&+ \big(\sum^{k-1}_{i=0}g_{\text{val}}^{(i+1)}g^{(i)}_{\text{soft}}(x_j)\big)^T \dfrac{\partial w_j^{\text{soft}}}{\partial \phi} \big]
\label{eq:metagradient}
\end{align}
Where  $g_{\text{val}}^{(i+1)}$ denotes the gradient of student parameters on the validation set at the $(i+1)$-th inner loop. $g^{(i)}_{\text{hard}}(x_j)$ and $g^{(i)}_{\text{soft}}(x_j)$ denote the student gradients with respect to the hard loss and the soft loss, respectively, computed on sample $x_j$ at inner step $i$. 
For each sample $x_j$, if the information accumulated over $k$ inner steps is aligned with the validation gradient, 
\[
\sum_{i=0}^{k-1} g_{\text{val}}^{(i+1)}\, g_{\text{hard}}^{(i)}(x_j) > 0,
\]
then the meta update at time step $t{+}1$ increases $w_j^{\text{hard}}$.
In other words, the meta model assigns larger hard-loss weights to samples whose gradients are directionally consistent with the balanced validation signal. An analogous phenomenon holds for the soft term: samples assigned a large $w^{\text{soft}}$ must exhibit soft loss gradients aligned with the validation gradients.
As pointed out by Safaryan et al.~\cite{Safaryan2023KnowledgeDP}, the optimal distillation weight $\alpha^*$ increases with the expected alignment between the hard loss gradient at the student’s optimum and the loss gradient at the teacher’s parameters. In our setting, we instantiate the “optimum” direction for each sample $x_j$ by the validation gradient.
Consequently, given the loss pair $[\mathrm{CE}(y_i, p_T( x_i)), \mathrm{CE}(y_i, p_S( x_i))]$, the meta model learns to adjust the weights so that the
resulting student aligns with the balanced set, implicitly steering learning toward better tail classes generalization.
\section{Experiments}
\label{sec:experiments}

\subsection{Settings}

\textbf{Datasets.}
We evaluate our method on the CIFAR long-tailed~\cite{imbalancedcifar}, which is widely used to assess performance under class imbalance.
The number of training samples per class is controlled by an exponential function, resulting in a class distribution defined as
$n_i = n_{\max} \times \mu^{\frac{i-1}{C-1}}$, 
where \( n_i \) is the number of samples in class \( i \), \( n_{\max} \) is the number of samples in the largest class, \( C \) is the total number of classes, and \( \mu \in (0,1] \) controls the imbalance severity.
The \emph{imbalance factor ($\rho$)} is defined as
\(\text{$\rho$} = \dfrac{n_{\max}}{n_{\min}}\). We use only 1000 images as validation set of unbiased data examples for both CIFAR-10 and CIFAR-100.\\
\textbf{Baselines.}
We compare our method with the following baselines:
\textbf{KD}~\cite{KD} - vanilla knowledge distillation with a fixed trade-off between hard and soft losses;
\textbf{WSL}\cite{WSL} - sample wise re-weighting of the KD loss by adapting the hard–soft balance per sample;
\textbf{RWKD}~\cite{RWKD} - meta learning based sample-wise re-weighting that adjusts KD using validation gradients;
\textbf{BKD}~\cite{BKD} - Balanced Knowledge Distillation with class-balanced weights in the distillation term to better support minority classes;
\textbf{CCKD}~\cite{CCKD}- Class-Conditional Knowledge Distillation that matches class-conditional distributions to transfer inter-class relational knowledge;
\textbf{LFME}~\cite{LFME}- multi-expert KD that learns from multiple teachers with an adaptive teacher-selection strategy;
\textbf{DiVE}~\cite{DiVE}- distillation via virtual examples that flatten the teacher’s effective label distribution for long-tailed data.\\
\textbf{Models.} We use ResNet-32x4, ResNet-8x4~\cite{resnet} for teacher and student backbones, respectively.
The meta network is implemented as a three-layer MLP with two hidden layers (Tanh activation) and sigmoid outputs for sample-wise weight generation. We train the teacher and student model using SGD with momentum 0.9 and weight decay $5 \times 10^{-4}$. 
The initial learning rate is set to $0.1$ and decayed by a factor of 10 at the 80-th and 100-th epoch, with a total of 120 epochs. 
The meta network is optimized using Adam with a learning rate of $1 \times 10^{-3}$. We use temperature $\tau = 4$ for knowledge distillation. 
\begin{table}
\centering
\caption{Test accuracy (\%) on datasets with imbalance factors $\rho$. The table runs over three seeds and takes the average.}
\label{tab:main_results}
\begin{tabular}{c|cccc|cccc}
\toprule
Dataset & \multicolumn{4}{c|}{Long-Tailed CIFAR-10} & \multicolumn{4}{c}{Long-Tailed CIFAR-100} \\
Imbalance & $\rho=1$ & $\rho=10$ & $\rho=50$ & $\rho=100$  & $\rho=1$ & $\rho=10$ & $\rho=50$ & $\rho=100$ \\
\midrule
Teacher(CE) & 95.82 & 90.79 & 82.72 & 77.23 & 79.4 & 65.45 & 51.59 & 45.88 \\
Student(CE) & 92.63 & 87.41 & 78.32 & 71.0 & 72.41 & 59.97 & 48.14 & 42.42 \\
\midrule
KD~\cite{KD}      & 93.15 & 87.46 & 77.27 & 70.11 & 73.04 & 59.38 & 47.87 & 42.91 \\
BKD~\cite{BKD}    & 92.58 & 87.03 & \underline{81.67} & \textbf{77.73} & 73.12 & \underline{62.38} & 50.71 & \underline{46.07} \\
CCKD~\cite{CCKD}  & 92.96 & 87.61 & 79.74 & 73.82 & 72.91 & 62.11 & 48.86 & 44.92 \\
WSL~\cite{WSL} & \textbf{93.91} & 87.5 & 76.87 & 68.53 & \textbf{75.18} & 61.56 & 47.2 & 42.26 \\
RWKD~\cite{RWKD} & 93.25 & 88.08 & 78.97 & 72.78 & 73.68 & 61.38 & 47.26 & 42.83 \\
LFME~\cite{LFME} & 92.85 & 88.07 & 80.28 & 74.04 & 73.13 & 61.51 & 49.29 & 45.18 \\
DiVE~\cite{DiVE} & 91.55 & \underline{88.2} & 80.79 & 74.83 & 71.89 & 61.93 & \underline{51.01} & 44.76 \\
\midrule
\textbf{Ours} & \underline{93.85} & \textbf{88.85} & \textbf{82.45} & \underline{75.91} & \underline{74.08} & \textbf{64.69} & \textbf{51.84 }&\textbf{ 46.23} \\
\bottomrule
\end{tabular}
\end{table}
\subsection{Results and ablation studies}

Table~\ref{tab:main_results} reports test accuracy on CIFAR long-tailed datasets for different imbalance factors $\rho$. On long-tailed CIFAR-10, BiKD matches or surpasses the strongest balanced KD baselines in most imbalance regimes. In particular, at $\rho = 50$, \textbf{BiKD} achieves 82.45\% accuracy, outperforming \textbf{BKD} (81.67\%), \textbf{DiVE} (80.79\%), and \textbf{LFME} (80.28\%). On long-tailed CIFAR-100, \textbf{BiKD} yields consistent gains over \textbf{BKD} at all imbalance levels, suggesting that sample-wise bilevel weighting scales well to more classes.

\begin{figure}[t]
    \centering
    \begin{subfigure}[b]{0.66\linewidth}
        \centering
        \includegraphics[width=\linewidth]{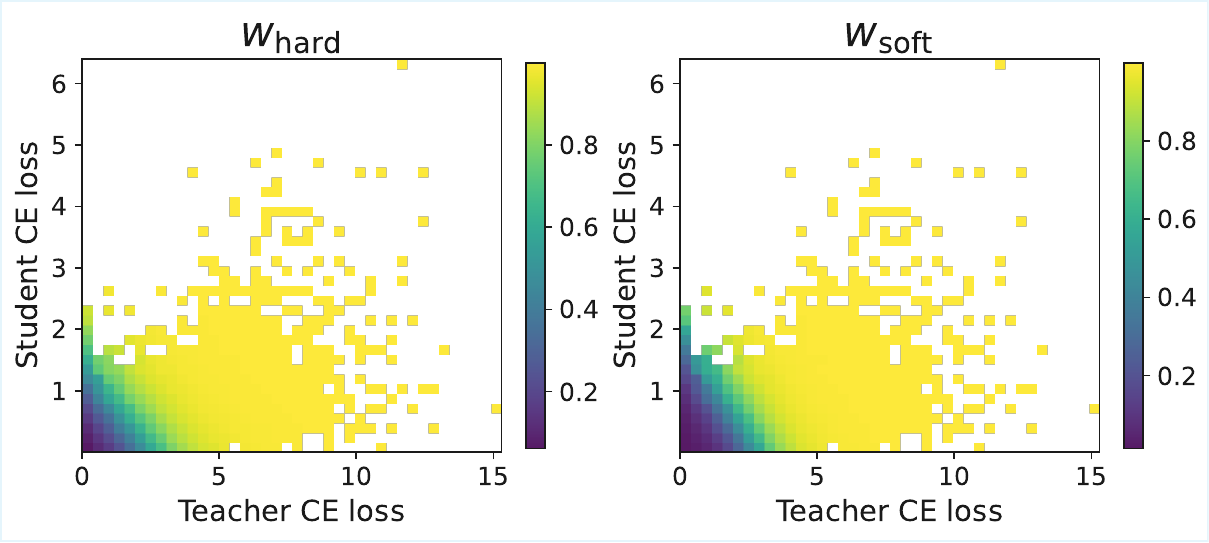}
        \caption{}
        \label{fig:weight_visualization}
    \end{subfigure}
    \hfill
    \begin{subfigure}[b]{0.33\linewidth}
        \centering
        \includegraphics[width=\linewidth]{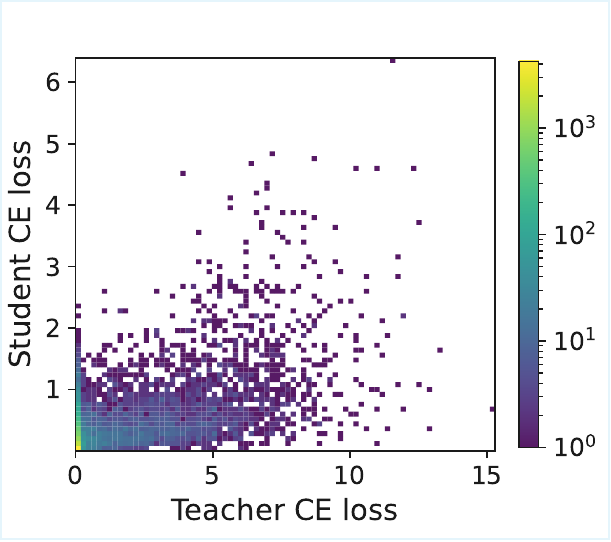}
        \caption{}
        \label{fig:density_training_data}
    \end{subfigure}

    \caption{Visualization of the meta outputs after training process with long-tailed CIFAR-100($\rho=50$). Each point corresponds to a training sample, with color indicating the value of assigned weight.}
\end{figure}

Figure \ref{fig:weight_visualization} visualizes outputs of the weight model on the long-tailed CIFAR-100 with $\rho = 50$, where each point corresponds to a training sample after the training process. Each point is located by its teacher and student cross entropy losses $\mathrm{CE}(y_i, p_T( x_i)), \mathrm{CE}(y_i, p_S( x_i))$ and colored by the assigned weight. The two panels reveal a behavior of the meta model: samples with small losses $\mathrm{CE}(y_i, p_T( x_i)), \mathrm{CE}(y_i, p_S( x_i))$ are assigned small weights $w_i^{\text{soft}}$ and $w_i^{\text{hard}}$. This prevents easy examples (head classes) from dominating training and harming the student generalization. In contrast, samples whose loss pairs remain large, which is typical for tail-class examples, tend to receive larger hard and soft weights. Therefore, the student could focus more on training these samples. Figure \ref{fig:density_training_data} demonstrates the density of 50000 training samples after the training process. This scatter plot shows that the teacher exhibits a strong bias on a small subset of the data, and its loss fails to converge well for many samples. In contrast, the student cross entropy $\mathrm{CE}(y_i, p_S(x_i))$ converges better, leaving only a small fraction of samples with large loss values. This indicates that the student learns more stably than the teacher, which explains why it eventually surpasses the teacher in performance, as reported in Table~\ref{tab:main_results}.

\begin{figure}[t]
    \centering
    \begin{subfigure}[b]{0.66\linewidth}
        \centering
        \includegraphics[width=\linewidth]{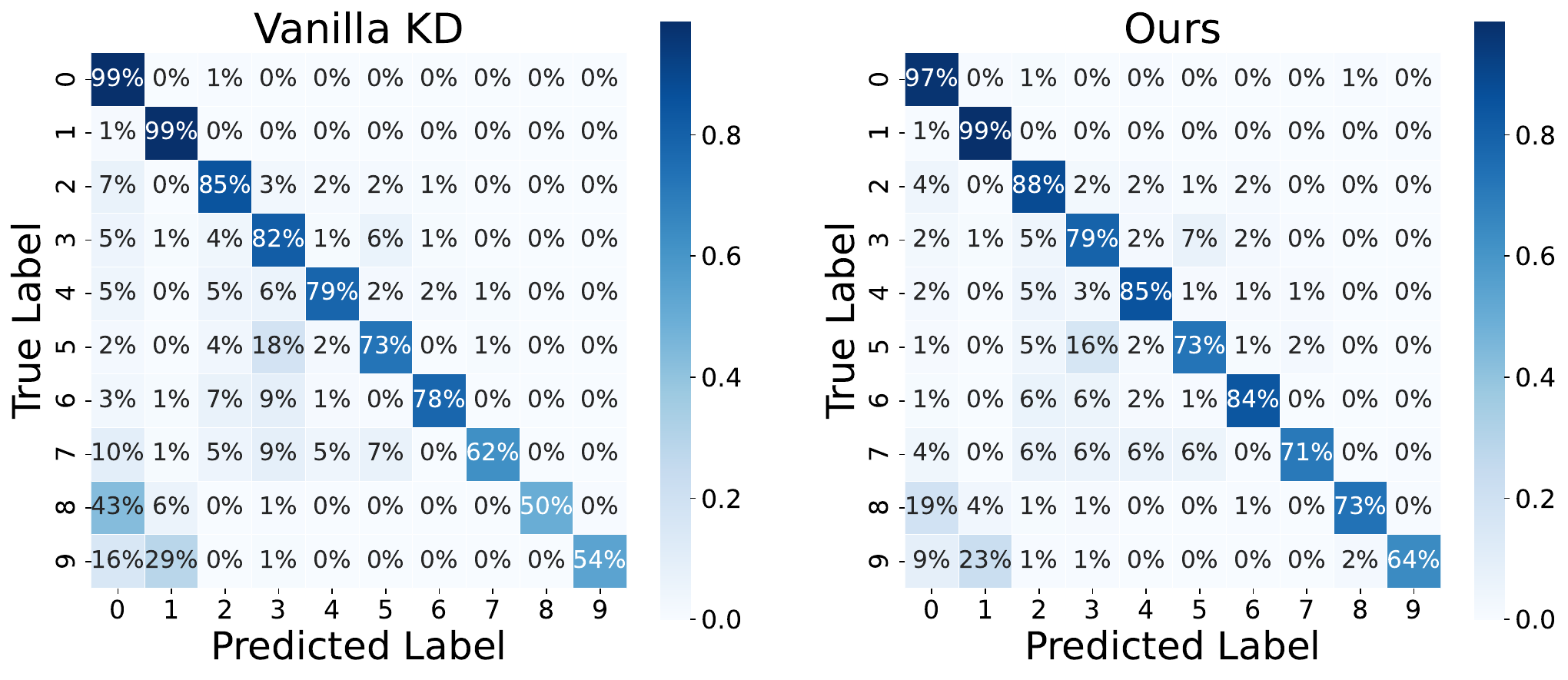}
        \caption{}
        \label{fig:confusion_matrix}
    \end{subfigure}
    \hfill
    \begin{subfigure}[b]{0.33\linewidth}
        \centering
        \includegraphics[width=\linewidth,height=0.85\linewidth]{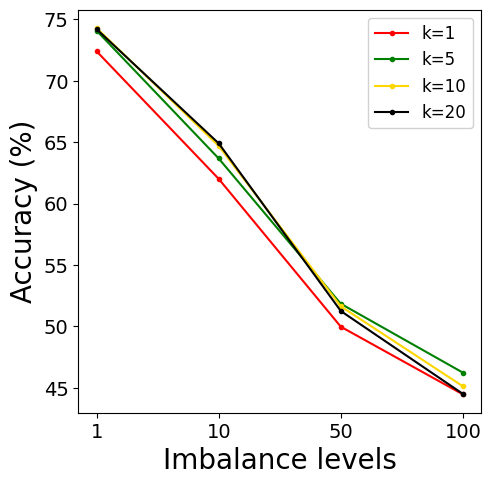}
        \caption{}
        \label{fig:inner_steps}
    \end{subfigure}

    \caption{In the left subfigure, the confusion matrix for the Vanilla KD and ours on long-tailed CIFAR-10 ($\rho=50$) are plotted. In the right figure, we show the comparison between different inner steps k on long-tailed CIFAR-100.}
\end{figure}

Figure \ref{fig:confusion_matrix} further visualizes confusion matrices on long-tailed CIFAR-10 ($\rho=50$). Compared to vanilla \textbf{KD}, \textbf{BiKD} dramatically improves the accuracy for tail classes. This supports our claim that bilevel sample-wise weighting generalizes more on tail classes, while still preserving strong performance on majority classes. In the subfigure \ref{fig:inner_steps}, we empirically show that $k=5$ is preferable for the best accuracy for all settings. While other reweight methods~\cite{RWKD,Ren2018LearningTR,Shu2019MetaWeightNetLA} estimate the optimized $\theta^*(\phi)$ using an approximation with an inner loop step, our proposed BiKD uses multiple SGD updates to mitigate the noise in the estimation of $\theta^*(\phi)$, leading to the best performance of the student model.
\section{Conclusion}
In this paper, we introduced BiKD, a bilevel knowledge distillation framework for imbalanced datasets. Within this framework, a small weight network produces per-sample weights for the hard and soft terms. This design allows the student to use an unconstrained adaptive combination of hard and soft losses, mitigating the bias induced by majority classes. We further proposed a multi-step SGD strategy without unrolling full computation graphs. Experiments on long-tailed CIFAR-10/100 demonstrate that BiKD surpasses recent balanced distillation methods across a range of imbalance factors. In addition, theoretical and experimental analyses of the learned weights reveal that the student model mainly focuses on hard tail samples, explaining the gains in tail accuracy.

\section{Appendix: Derivation of meta gradients}
We show how to compute exactly the gradient of the meta update gradient at time step $t$. By using implicit differentiation, we have:
\begin{equation}
\nabla_\phi = \sum_{i=0}^{k-1} \nabla_\phi \mathcal{L}_{\text{val}}(\theta^{(t+i+1)}) = \sum_{i=0}^{k-1} \nabla_\theta \mathcal{L}_{\text{val}}(\theta^{(t+i+1)}) \nabla_\phi \theta^{(t+i+1)} =\sum_{i=0}^{k-1} g_{\text{val}}^{(i+1)}\nabla_\phi \theta^{(t+i+1)}
\end{equation}
Where  $g_{\text{val}}^{(i+1)}$ denotes the gradient of student parameters on the validation set at the $(i+1)$-th inner loops. 
From \eqref{eq:student_update} in which $\mathcal{L}_{\text{train}}$ is computed on the training set, we obtain $\nabla_\phi \theta^{(t+i+1)} $ as follows:
\begin{align}
\nabla_\phi \theta^{(t+i+1)} &= \nabla_\phi\big(\theta^{(t+i)} - \eta_\theta \nabla_\theta\mathcal{L}_{\text{train}}(\theta^{(t+i)},\phi^{(t)}) \big) \label{5}\\
&=-\eta_\theta \nabla_\phi \nabla_\theta \frac{1}{N} \sum_{x_j \in \mathcal{D}_{\text{train}}}\left[ w_j^{\text{hard}}\ \mathcal{L}^{(i)}_{\text{hard}}(x_j) + w_j^{\text{soft}} \mathcal{L}^{(i)}_{\text{soft}}(x_j) \right ] \label{6}\\
&= -\dfrac{\eta_\theta}{N} \sum_{x_j \in \mathcal{D}_{\text{train}}} \left[ \dfrac{\partial \mathcal{L}^{(i)}_{\text{hard}}(x_j)}{\partial \theta}^T \dfrac{\partial w_j^{\text{hard}}}{\partial \phi} + \dfrac{\partial \mathcal{L}^{(i)}_{\text{soft}}(x_j)}{\partial \theta}^T \dfrac{\partial w_j^{\text{soft}}}{\partial \phi}  \right] \notag \\
&= -\dfrac{\eta_\theta}{N} \sum_{x_j \in \mathcal{D}_{\text{train}}} \left[ g^{(i)}_{\text{hard}}(x_j)^T \dfrac{\partial w_j^{\text{hard}}}{\partial \phi} + g^{(i)}_{\text{soft}}(x_j)^T \dfrac{\partial w_j^{\text{soft}}}{\partial \phi}  \right] \notag
\end{align}
The equality from \eqref{5} to \eqref{6} follows because our method does not unroll the student’s computation graph through the inner loop. We treat $\theta$ as detached with respect to $\phi$ at each inner step, hence $\theta^{(t+i)}$ is independent of $\phi$, and the term $\nabla_\phi \theta^{(t+i)}$ equals to 0.\\
Therefore, we obtain the one step hypergradient for the $(i+1)$-th inner update at time step $t$ as:\\
\[
\nabla_\phi \mathcal{L}_{\text{val}}(\theta^{(t+i+1)}) = -\dfrac{\eta_\theta}{N} \sum_{x_j \in \mathcal{D}_{\text{train}}}\left[\big(g_{\text{val}}^{(i+1)} g^{(i)}_{\text{hard}}(x_j)\big)^T \dfrac{\partial  w_j^{\text{hard}}}{\partial \phi} + \big(g_{\text{val}}^{(i+1)}g^{(i)}_{\text{soft}}(x_j)\big)^T \dfrac{\partial w_j^{\text{soft}}}{\partial \phi}  \right] 
\]
Then the gradient which used to update the meta model can be computed by:
\begin{align*}
\nabla_\phi &= \sum_{i=0}^{k-1} \nabla_\phi \mathcal{L}_{\text{val}}(\theta^{(t+i+1)})\\
&= -\dfrac{\eta_\theta}{N} \sum_{x_j \in \mathcal{D}_{\text{train}}} \left[\big( \sum^{k-1}_{i=0}g_{\text{val}}^{(i+1)} g^{(i)}_{\text{hard}}(x_j)\big)^T \dfrac{\partial w_j^{\text{hard}}}{\partial \phi} + \big(\sum^{k-1}_{i=0}g_{\text{val}}^{(i+1)}g^{(i)}_{\text{soft}}(x_j)\big)^T \dfrac{\partial w_j^{\text{soft}}}{\partial \phi}  \right]
\end{align*}
Combine the update formula of $\phi^{(t+1)}$, we have the equality in Eq.(\ref{eq:metagradient}).
\bibliographystyle{splncs04}
\bibliography{references}  

@article{KD,
  title={Distilling the Knowledge in a Neural Network},
  author={Hinton, Geoffrey E. and Vinyals, Oriol and Dean, Jeffrey},
  journal={arXiv},
  year={2015}
}

@article{BKD,
title = {Balanced knowledge distillation for long-tailed learning},
journal = {Neurocomputing},
year = {2023},
author = {Shaoyu Zhang and Chen Chen and Xiyuan Hu and Silong Peng},
}

@inproceedings{LFME,
  title={Learning From Multiple Experts: Self-paced Knowledge Distillation for Long-tailed Classification},
  author={Liuyu Xiang and Guiguang Ding},
  booktitle={ECCV},
  year={2020},
}

@article{SHIKE,
  title={Long-Tailed Visual Recognition via Self-Heterogeneous Integration with Knowledge Excavation},
  author={Yang Jin and Mengke Li and Yang Lu and Yiu-ming Cheung and Hanzi Wang},
  journal={CVPR},
  year={2023},
}

@article{he2024joint,
  title={Joint weighted knowledge distillation and multi-scale feature distillation for long-tailed recognition},
  author={He, Yiru and Wang, Shiqian and Yu, Junyang and Liu, Chaoyang and He, Xin and Li, Han},
  journal={International Journal of Machine Learning and Cybernetics},
  year={2024},
}

@article{DiVE,
  title={Distilling Virtual Examples for Long-tailed Recognition},
  author={Yingying He and Jianxin Wu and Xiu-Shen Wei},
  journal={ICCV},
  year={2021},
}

@article{Safaryan2023KnowledgeDP,
  title={Knowledge Distillation Performs Partial Variance Reduction},
  author={Mher Safaryan and Alexandra Peste and Dan Alistarh},
  journal={ArXiv},
  year={2023},
}

@inproceedings{CCKD,
  title={Enhancing Class-Imbalanced Learning with Pre-Trained Guidance through Class-Conditional Knowledge Distillation},
  author={Lan Li and Xin-Chun Li and Han-Jia Ye and De-Chuan Zhan},
  booktitle={ICML},
  year={2024},
}

@inproceedings{WSL,
  title={Rethinking Soft Labels for Knowledge Distillation: A Bias-Variance Tradeoff Perspective},
  author={Helong Zhou and Liangchen Song and Jiajie Chen and Ye Zhou and Guoli Wang and Junsong Yuan and Qian Zhang},
  booktitle={ICLR},
  year={2021},
}

@inproceedings{RWKD,
  title={RW-KD: Sample-wise Loss Terms Re-Weighting for Knowledge Distillation},
  author={Peng Lu and Abbas Ghaddar and Ahmad Rashid and Mehdi and Ali Ghodsi and Philippe Langlais},
  booktitle={EMNLP},
  year={2021},
}

@inproceedings{Ren2018LearningTR,
  title={Learning to Reweight Examples for Robust Deep Learning},
  author={Mengye Ren and Wenyuan Zeng and Binh Yang and Raquel Urtasun},
  booktitle={ICML},
  year={2018},
}

@inproceedings{Shu2019MetaWeightNetLA,
  title={Meta-Weight-Net: Learning an Explicit Mapping For Sample Weighting},
  author={Jun Shu and Qi Xie and Lixuan Yi and Qian Zhao and Sanping Zhou and Zongben Xu and Deyu Meng},
  booktitle={Neurips},
  year={2019},
}

@article{survey_imbalanced,
  title={Deep Long-Tailed Learning: A Survey},
  author={Yifan Zhang and Bingyi Kang and Bryan Hooi and Shuicheng Yan and Jiashi Feng},
  journal={IEEE Transactions on Pattern Analysis and Machine Intelligence},
  year={2021},
}

@article{FitNet,
  title={FitNets: Hints for Thin Deep Nets},
  author={Adriana Romero and Nicolas Ballas and Samira Ebrahimi Kahou and Antoine Chassang and Carlo Gatta and Yoshua Bengio},
  journal={CoRR},
  year={2014},
}

@article{Zagoruyko2016PayingMA,
  title={Paying More Attention to Attention: Improving the Performance of Convolutional Neural Networks via Attention Transfer},
  author={Sergey Zagoruyko and Nikos Komodakis},
  journal={ArXiv},
  year={2016},
}

@article{Park2019RelationalKD,
  title={Relational Knowledge Distillation},
  author={Wonpyo Park and Dongju Kim and Yan Lu and Minsu Cho},
  journal={CVPR},
  year={2019},
}

@article{Liu2019LargeScaleLR,
  title={Large-Scale Long-Tailed Recognition in an Open World},
  author={Ziwei Liu and Zhongqi Miao and Xiaohang Zhan and Jiayun Wang and Boqing Gong and Stella X. Yu},
  journal={CVPR},
  year={2019},
}

@inproceedings{CBD,
  title={Class-Balanced Distillation for Long-Tailed Visual Recognition},
  author={Ahmet Iscen and Andre F. de Ara{\'u}jo and Boqing Gong and Cordelia Schmid},
  booktitle={British Machine Vision Conference},
  year={2021},
}

@article{Heo2019ACO,
  title={A Comprehensive Overhaul of Feature Distillation},
  author={Byeongho Heo and Jeesoo Kim and Sangdoo Yun and Hyojin Park and Nojun Kwak and Jin Young Choi},
  journal={ICCV},
  year={2019},
}

@article{guo2023classattentiontransferbased,
      title={Class Attention Transfer Based Knowledge Distillation}, 
      author={Ziyao Guo and Haonan Yan and Hui Li and Xiaodong Lin},
      journal={CVPR},      
      year={2023},
}

@article{Tung2019SimilarityPreservingKD,
  title={Similarity-Preserving Knowledge Distillation},
  author={Frederick Tung and Greg Mori},
  journal={ICCV},
  year={2019},
}

@article{KDsurvey,
  title={Knowledge Distillation: A Survey},
  author={Jianping Gou and B. Yu and Stephen J. Maybank and Dacheng Tao},
  journal={International Journal of Computer Vision},
  year={2020},
}

@article{imbalancedcifar,
  title={Class-Balanced Loss Based on Effective Number of Samples},
  author={Yin Cui and Menglin Jia and Tsung-Yi Lin and Yang Song and Serge J. Belongie},
  journal={CVPR},
  year={2019},
}

@article{resnet,
  title={Deep Residual Learning for Image Recognition},
  author={Kaiming He and X. Zhang and Shaoqing Ren and Jian Sun},
  journal={CVPR},
  year={2015},
}
\end{document}